\documentclass[letterpaper, 10 pt, journal, twoside]{IEEEtran}
\usepackage[sort,compress]{cite}
\usepackage{stackengine}
\usepackage{textpos}
\usepackage{common}
\usepackage{this_doc}
\usepackage{lipsum} 
\hypersetup{colorlinks,urlcolor=blue}
\newcommand{\highlight}[1]{{\color{black}#1}}
\newcommand{\highlightCamReady}[1]{{\color{black}#1}}

\begin{document}
\title{
Optimizing Fiducial Marker Placement for Improved Visual Localization
}

\author{Qiangqiang Huang$^{1}$,
Joseph DeGol$^{2}$, Victor Fragoso$^{2}$, Sudipta N. Sinha$^{2}$, and John J. Leonard$^{1}$
\thanks{$^{1}$Computer Science and Artificial Intelligence Lab,
Massachusetts Institute of Technology, Cambridge, MA 02139, USA{\tt\small \{hqq, jleonard\}@mit.edu}.
}%
\thanks{$^{2}$Microsoft, Redmond, WA 98052, USA{\tt\small \{jodegol, victor.fragoso, sudipta.sinha\}@microsoft.com}.
}%
\thanks{The code is available at \url{https://github.com/doublestrong/OMP}. This work was started during Qiangqiang's internship at Microsoft and extended at MIT. Qiangqiang and John were partially supported by ONR grant N00014-18-1-2832 and ONR Neuroautonomy MURI grant N00014-19-1-2571.
}%
\vspace{-1cm}
}

\markboth{}
{Huang \MakeLowercase{\textit{et al.}}: Optimizing Fiducial Marker Placement for Improved Visual Localization} 
\maketitle

\begin{textblock*}{0.9\textwidth}(.05\textwidth,-5.55cm)
  \begin{center}
    This is an extended technical report for our publication in the \emph{IEEE
      Robotics and
      Automation Letters}. \\
    Please cite the paper as: Q. Huang, J. DeGol, V. Fragoso, S. N. Sinha, and J. J. Leonard, \\
    ``Optimizing Fiducial Marker Placement for Improved Visual Localization'',\\ \emph{IEEE Robotics and
      Automation Letters (RA-L)}, 2023.
  \end{center}
\end{textblock*}

\begin{abstract}
Adding fiducial markers to a scene is a well-known strategy for making visual localization algorithms more robust. Traditionally, these marker locations are selected by humans who are familiar with visual localization techniques. This paper explores the problem of automatic marker placement within a scene. Specifically, given a predetermined set of markers and a scene model, we compute optimized marker positions within the scene that can improve accuracy in visual localization. Our main contribution is a novel framework for modeling camera localizability that incorporates both natural scene features and artificial fiducial markers added to the scene. We present optimized marker placement (OMP), a greedy algorithm that is based on the camera localizability framework. We have also designed a simulation framework for testing marker placement algorithms on 3D models and images generated from synthetic scenes. We have evaluated OMP within this testbed and demonstrate an improvement in the localization rate by up to 20 percent on four different scenes.
\end{abstract}
\begin{IEEEkeywords}
Localization, Computer Vision for Automation, Landmark Deployment, Fiducial Markers.
\end{IEEEkeywords}
\IEEEpeerreviewmaketitle

\section{Introduction}
\begin{figure}[!t]
    \centering
    \includegraphics[width=.8\linewidth]{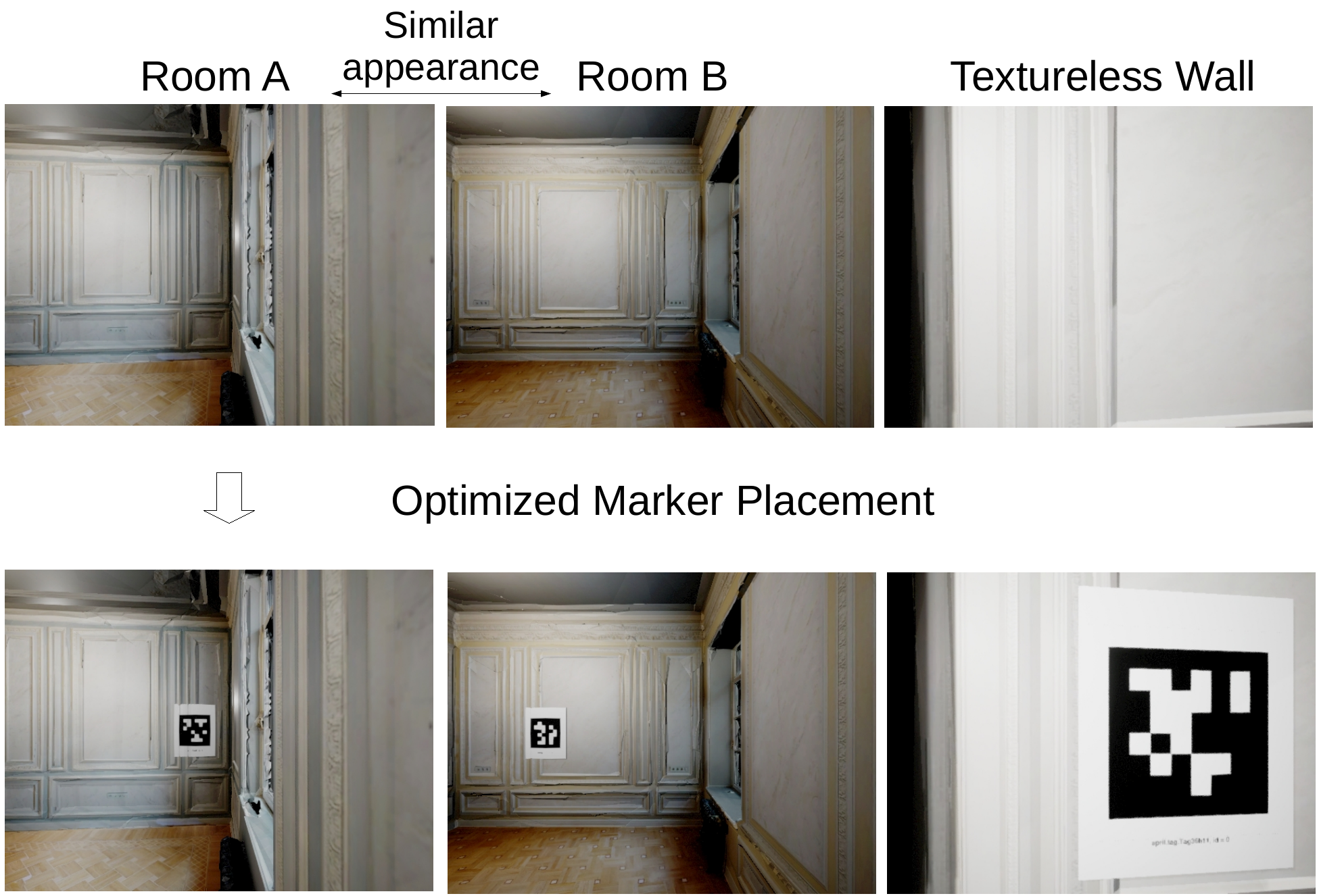}
    \caption{Three challenging examples for visual localization. The images on the left and middle show two almost identical rooms in the scene, whereas the image on the right depicts a very weakly textured surface. Marker placements\protect\footnotemark in this scene guided by our optimized marker placement approach led to improved visual localization on these examples.
    }
    \label{fig:demo-challenges}
\end{figure}

\IEEEPARstart{V}{isual} localization is a foundational technique for AR/VR, autonomous driving, and robotic navigation and manipulation. A typical problem in visual localization is to estimate the camera pose of a query image, provided a pre-built map. While the problem has long been investigated in many fields~\cite{zhang2021reference}, visual localization still suffers due to challenging scenes such as textureless walls and repetitive structures (e.g., Rooms A and B in Fig.~\ref{fig:demo-challenges}). One common solution to these challenges is to place fiducial markers as additional texture and identifiers in the scene~\cite{munoz2020ucoslam,DeGol_2018_ECCV}; however, placing fiducial markers in larger environments is a time consuming process and the resulting performance improvement depends on marker positions. Thus, optimizing marker placement is valuable for robust visual localization.

This work proposes an automatic approach to optimizing marker placement such that 1) the resulting marker positions yield improved accuracy in visual localization and 2) a human user will be able to place markers at positions planned by the approach (e.g., no markers on the ceiling). Specifically, the approach computes optimized marker positions, given a predetermined set of markers and a scene model. The key contributions of this work include:
\begin{enumerate}
    \item This is the first work that optimizes marker placement for visual localization based on scene features and fiducial markers.
    \item We propose a novel framework that models localizability of camera poses in a scene and computes localizability scores.
    \item We develop a greedy algorithm that optimizes marker positions with the goal of increased localizability scores.
    \item We design a simulation framework for testing marker placement algorithms on 3D scene models that enables others to reproduce and build on our work.
    \item We demonstrate that optimized marker placement by our approach can improve the localization rate by up to 20 percent on four different scenes.
\end{enumerate}

\footnotetext{Fiducial markers in the examples are AprilTags\cite{olson2011apriltag} but our algorithm is general and can be used with any existing family of fiducial markers.}

\section{Related Work}
We briefly review some recent work related to mapping and localization with fiducial markers and marker/landmark placement optimization. Examples of fiducial markers include tag families with explicit IDs (e.g., ArUco markers~\cite{aruco2014}, AprilTag~\cite{olson2011apriltag}, ChromaTag~\cite{DeGol:ICCV:17}) and emerging learning-based marker designs~\cite{zhang2022deeptag}. Fiducial markers are widely recognized as an effective approach for improving localization and mapping accuracy. DeGol et al.~\cite{DeGol_2018_ECCV} demonstrate that marker IDs are useful in image matching and resectioning for structure from motion (SfM), leading to improvements in reconstruction results. The UcoSLAM system~\cite{munoz2020ucoslam} integrates marker detection with a bag-of-words approach and presents more robust tracking and relocalization than SLAM techniques with no marker detection~\cite{mur2017orb, gao2018ldso}. However, marker placements in these SfM or SLAM systems are manually determined and not planned by algorithms.

Existing work about marker deployment focuses on robotic localization without considering scene features \cite{chen2006practical,vitus2011sensor,jourdan2008optimal}. Beinhofer et al. \cite{beinhofer2013effective} explore optimal placement of artificial landmarks such that a robot equipped with range and/or bearing sensors repeatedly follows predetermined trajectories in planar environments with improved accuracy. \highlight{Meyer-Delius et al. \cite{meyer2011using} introduce a measure that defines the uniqueness of robot poses in the context of Monte Carlo localization using laser scanners and then propose a greedy algorithm to incrementally select landmark locations for maximizing the measure. While we find the greedy algorithm is similar to ours, it is not straightforward to apply the measure to visual localization using images and scene features}. Lei et al. \cite{lei2022tie} investigate landmark deployment for poses on $\SE(3)$ and demonstrate placing fiducial markers in a cubic environment; however, features in the scene are not involved in optimizing the marker placement.
\section{Methods}
\begin{figure}[!t]
    \centering
    \includegraphics[width=0.8\linewidth]{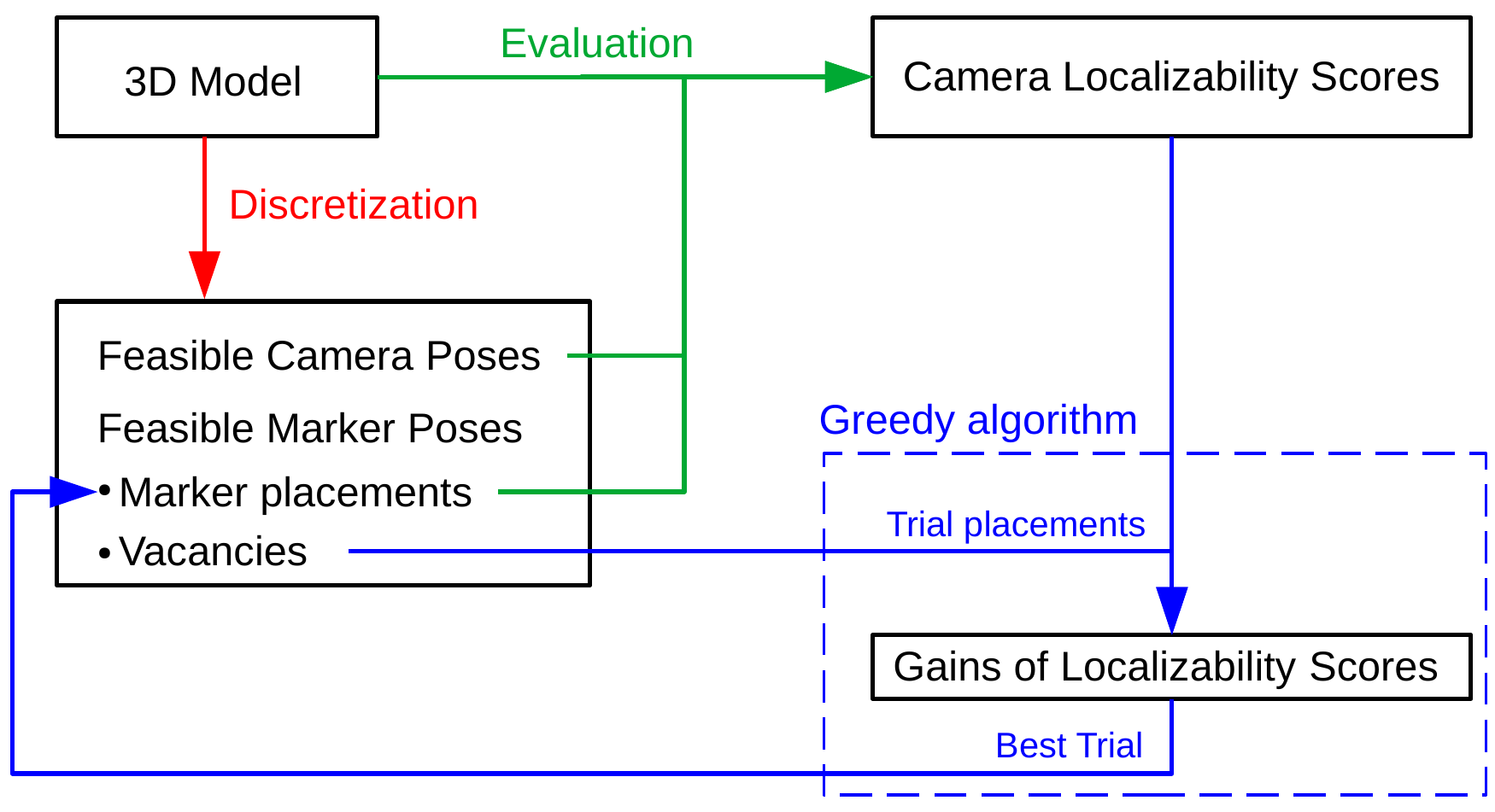}
    \caption{An overview of our approach. We first create a set of feasible camera poses and marker poses by discretizing space in the 3D model. Then we evaluate localizability scores of the feasible camera poses and update the scores once a feasible marker pose is selected to place a marker. The marker placement is selected by a greedy algorithm as the best trial out of trial placements in the vacancies (unselected marker poses). These trial placements are ranked by gains of localizability scores.}
    \label{fig:overview}
\end{figure}
We aim to compute $k$ 3D locations in the scene for placing $k$ fiducial markers
such that after marker placement, the camera localization performance improves
for query images from anywhere within the scene. \highlight{In summary, we solve the global search of optimal $k$ locations by a greedy algorithm that seeks one marker placement each time.}
\subsection{Assumptions}
This work makes two assumptions: 1) A textured 3D model of the scene is available, and 2) markers and cameras are located on a 3D plane parallel to the ground plane at roughly the eye level of a person with average height. Note that the textured model can be a 3D simulation environment or a dense reconstruction of scenes. We will collect images (e.g., RGB, depth, and surface normal) and corresponding camera poses from the model and take them as input to our approach for optimizing marker placement. The second assumption ensures that our marker placement will be reachable to a human user and constrains the number of feasible camera and marker locations for computational efficiency.
\subsection{Key Elements of Proposed Approach}
\label{sec:key-tech}
Fig.~\ref{fig:overview} shows an overview of our approach, which is composed of three key elements: 1) discretization, 2) evaluation of camera localizability, and 3) a greedy algorithm for selecting marker placements.

\begin{figure}[t!]
    \centering
    \includegraphics[width=.8\linewidth]{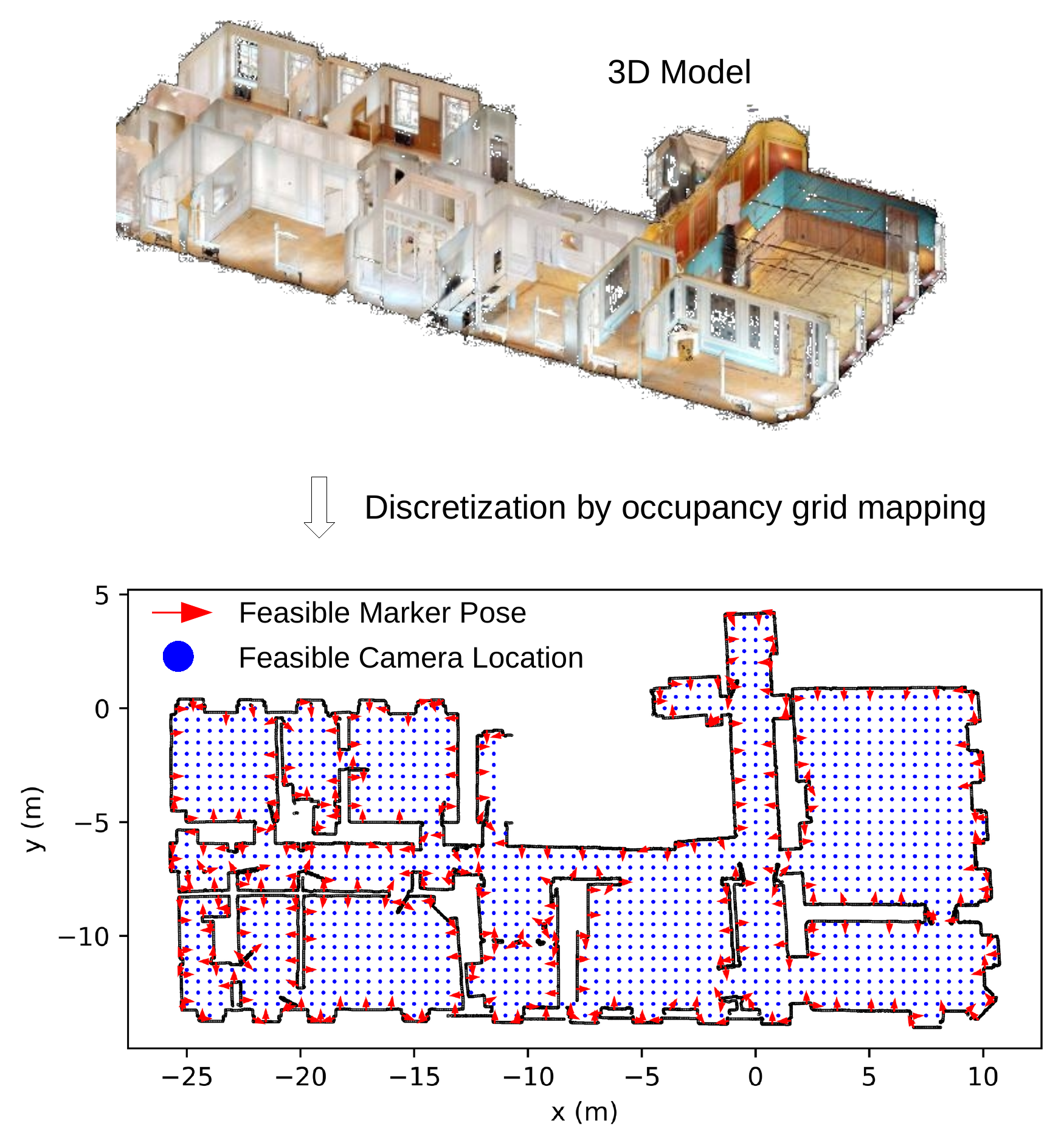}
    \caption{Discretization of a model from the Habitat-Matterport 3D dataset \cite{ramakrishnan2021hm3d}. We select a ground plane in the 3D model at roughly eye level of a human user. The discretized space of the ground plane consists of feasible marker poses (red arrows), which are sampled from scan points on the ground plane perimeter, and feasible camera locations (blue dots), which are centers of unoccupied cells in the 2D discrete grid.}
    \label{fig:discrete}
\end{figure}

\begin{figure*}[ht]
    \centering
    \includegraphics[width=.85\linewidth]{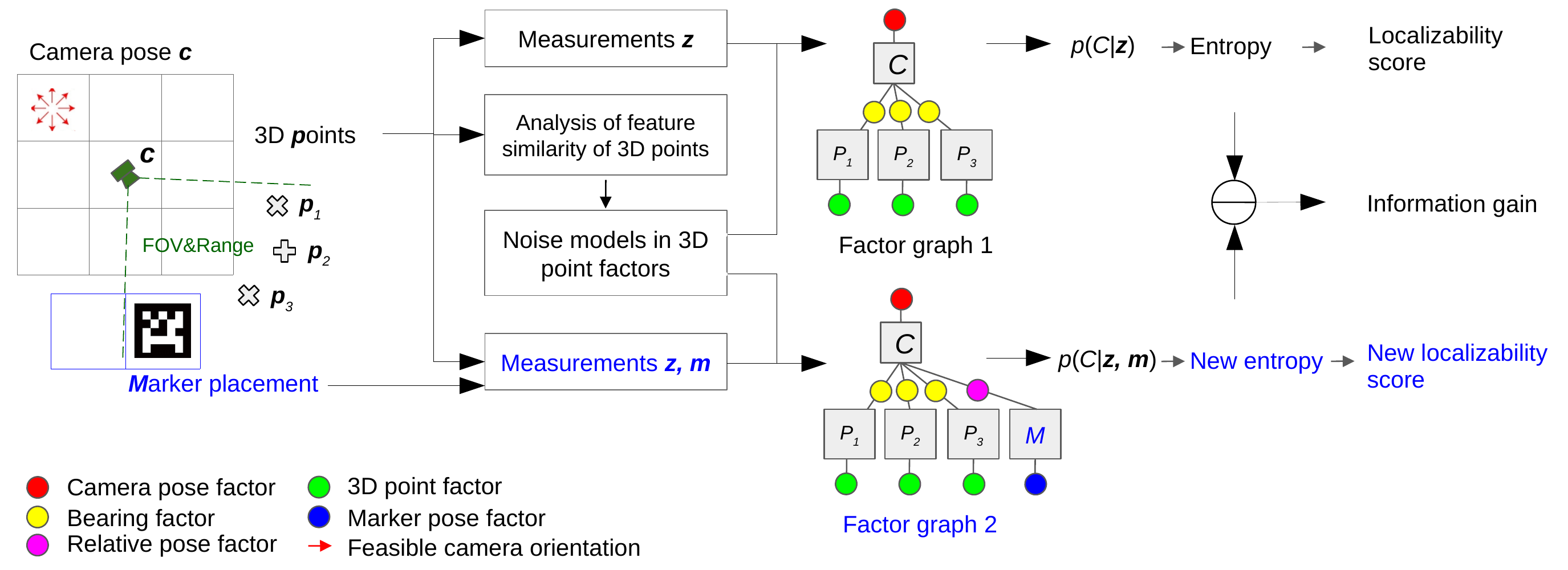}
    \caption{Evaluation of localizability scores and the information gain brought by a marker placement. On the left we show a grid of feasible camera poses. Feasible camera poses are positioned at cell centers with orientations shown as the red arrows. The field of view of camera pose $\camVal$ covers points $p_1$, $p_2$, and $p_3$ in the 3D model and a marker placement on the discretized perimeter of the level set of the ground plane. We synthesize measurements $\measVal$ of the points to create a camera localization problem using scene features. The problem is represented by factor graph 1 and distribution $p(\camVar|\measVal)$ by which we can compute the entropy as well as the localizability score of the camera pose seeing no markers. We penalize contributions of repetitive structures on the localizability score via the analysis of feature similarity. With additional measurements $\markerVal$ to the marker, we create another localization problem which is represented by factor graph 2 and distribution $p(\camVar|\measVal, \markerVal)$. The new problem leads to a new entropy and a new localizability score.}
    \label{fig:loc-score}
\end{figure*}

\begin{figure*}[h!]
    \centering
    \includegraphics[width=0.8\linewidth]{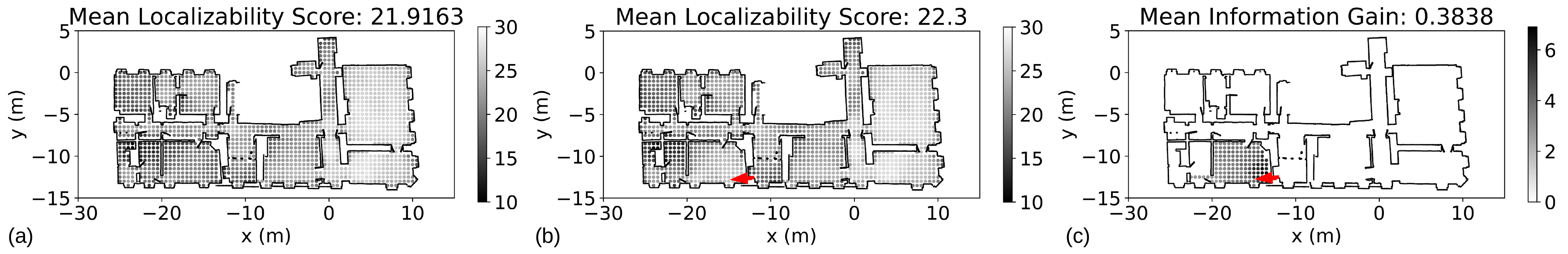}
    \caption{Results of localizability scores: (a) no markers, (b) a trial marker placement (red arrow), and (c) the information gain. The score (or gain) at a dot is the mean score (or gain) of camera poses at the dot with all feasible orientations. Darker dots stress low localizability scores in (a) and (b) and high information gains in (c). \highlight{This trial turns out to be the first marker in the optimized placement (see the Apartment in Fig.~\ref{fig:all-model-res}).}}
    \label{fig:info-gain-res}
\end{figure*}

\subsubsection{Discretization}We first convert the ground plane in the 3D model to a discretized space of camera and marker poses, as shown in Fig.~\ref{fig:discrete}. The conversion is implemented by occupancy grid mapping.
Centers of unoccupied grid cells are designated as feasible camera locations (dots in Fig.~\ref{fig:discrete}) while scan points form the perimeter of the free space (lines in Fig.~\ref{fig:discrete}). We uniformly downsample the scan points to generate a set of feasible marker poses $\markerSet$ (arrows in Fig.~\ref{fig:discrete}) whose orientations are determined by surface normals in the 3D model. \highlight{Note that one can choose other ways to select feasible marker poses and then still apply our marker placement algorithm. For example, the feasible marker poses can be further refined by incorporating semantics and physical constraints. It is possible that the algorithm could produce a marker placement in an infeasible location, although we found this was rare. Even so, we have done a sensitivity study showing that we can place the marker nearby the exact location and still get most of the gain\footnote{A sensitivity study about the influence of position and size deviations of markers on localization performance is available in Sec.\ref{sup:sensitivity}.}.} We derive a set of feasible camera poses $\camSet$ from the feasible camera locations. Each of the camera locations yields $n$ camera poses whose optical axes are parallel to the ground plane and evenly spaced in $[0, 2\pi]$ (e.g., the default $n=8$).

\subsubsection{Camera localizability score}We compute camera localizability scores by evaluating uncertainty in localizing feasible camera poses. Specifically, for any feasible camera pose $\camVal \in \camSet$ (the corresponding random variable is $\camVar$), we synthesize measurements $\measVal$ to create a camera localization problem, estimate the distribution of the camera pose $p(\camVar|\measVal)$, and define the localizability score of the camera pose $\locVal(\camVal)$ as the negation of the entropy of the distribution, as shown in
\begin{equation}
\locVal(\camVal) = -H(p(\camVar|\measVal))=\Expectation[\ln p(\camVar|\measVal)]. \label{eq:loc-score}
\end{equation}
If a new fiducial marker is added in the field of view (FOV) and \highlight{range} of the camera pose, the new synthetic measurement regarding the marker will change the entropy of the camera pose distribution, resulting in an information gain that quantifies the impact of the marker placement. Fig.~\ref{fig:loc-score} summarizes steps for evaluating the localizability score and the information gain. These steps are explained in detail in following paragraphs.

\textbf{Synthesized data for computing the localizability score}: The leftmost part of Fig.~\ref{fig:loc-score} illustrates 3D points and a feasible marker pose (i.e., trial marker placement) that are in the FOV\highlight{/range} of a feasible camera pose\footnote{\highlightCamReady{In practice, one can further refine marker poses in the FOV by considering marker sizes and rejecting corner cases that may fail the detection of markers. The cases include marker poses that are too close to the boundary of the view frustum of the camera.}}. We collect RGB and depth images at the camera pose in the 3D model. These images will be used to compute 3D points and descriptors of features (e.g., SIFT~\cite{Lowe04ijcv}). We use these known poses and points to synthesize measurements and estimate probability density functions (PDFs) of the camera pose variable. Measurements $\mathbf{z}$ in Fig.~\ref{fig:loc-score} contain the camera pose, the 3D points, and bearings between them. Thus the PDF $p(C|\mathbf{z})$, which is represented by factor graph 1, expresses the distribution of the camera pose constrained by the 3D points. Placing a marker in the FOV\highlight{/range} of the camera leads to new synthetic measurements $\mathbf{m}$ of the marker pose and the relative pose between the marker and the camera. As a result, the camera pose is further constrained by measurements $\mathbf{m}$ thus is described by a new PDF $p(C|\mathbf{z}, \mathbf{m})$ represented by factor graph 2 in Fig.~\ref{fig:loc-score}. We use an approach that is similar to the one proposed by Stachniss et al. \cite{stachniss2005information} to define the information gain of a marker placement. The information gain is defined as the change of entropy that the marker placement $\markerVal$ yields at the camera pose $\camVal$, as seen in
\begin{equation}
I(\markerVal,\camVal)=H(p(\camVar|\measVal))-H(p(\camVar|\measVal,\markerVal)).
\end{equation}

Fig.~\ref{fig:info-gain-res}a shows localizability scores of camera poses in the original ground plane with no marker placement. Note that the score at a dot in the figure is the mean score of camera poses with all feasible orientations. Fig.~\ref{fig:info-gain-res}b shows localizability scores after adding a marker (the arrow) to the ground plane perimeter. The scores increase in the region around the marker, indicated by the brighter dots in the region in Fig.~\ref{fig:info-gain-res}b and the information gain in Fig.~\ref{fig:info-gain-res}c.

\begin{algorithm}[t]
  \fontsize{8pt}{8pt}\selectfont
  \DontPrintSemicolon 
  \KwIn{The number of markers $k$, the list of feasible marker poses $\mathcal{M}$, the ground plane space $\mathcal{S}$}
  \KwOut{$k$ marker poses}
  \SetKwFor{RepTimes}{repeat}{times}{end}
  Initialize an empty list for storing selected marker poses $\mathcal{O}$
  
  \RepTimes{$k$}{
  	Initialize the best marker pose $T^{\star}=\emptyset$
  	
  	Initialize the highest localizability gain $g^{\star}=-\inf$

  	Evaluate localizability scores $\mathcal{L}^{\star}$ of camera poses in space $\mathcal{S}$
  	
    \For{Pose $T$ in $\mathcal{M}$}{
		Place a marker at pose $T$ in space $\mathcal{S}$
		
		Evaluate localizability scores $\mathcal{L}$ of camera poses \label{evaluate-scores}		

		Compute information gains $\mathcal{I}=\mathcal{L}-\mathcal{L}^{\star}$
				
		Evaluate localizability gain $g$ of the marker by \eqref{eq:marker-loc-gain} \label{evaluate-gains}
		
		\If{$g>g^{\star}$}{
			$T^{\star}=T$
			
			$g^{\star}=g$		
		}
		
		Remove the marker from space $\mathcal{S}$
    }
    Push $T^{\star}$ to $\mathcal{O}$

	Place a marker at pose $T^{\star}$ in space $\mathcal{S}$
    
    Remove $T^{\star}$ from $\mathcal{M}$
  }
  \Return{List of marker poses $\mathcal{O}$}\;
  \caption{Optimized Marker Placement (OMP)}
  \label{algo:imp}
\end{algorithm}
\setlength\floatsep{3pt}
\begin{figure}[h]
    \centering
    \includegraphics[width=0.9\linewidth]{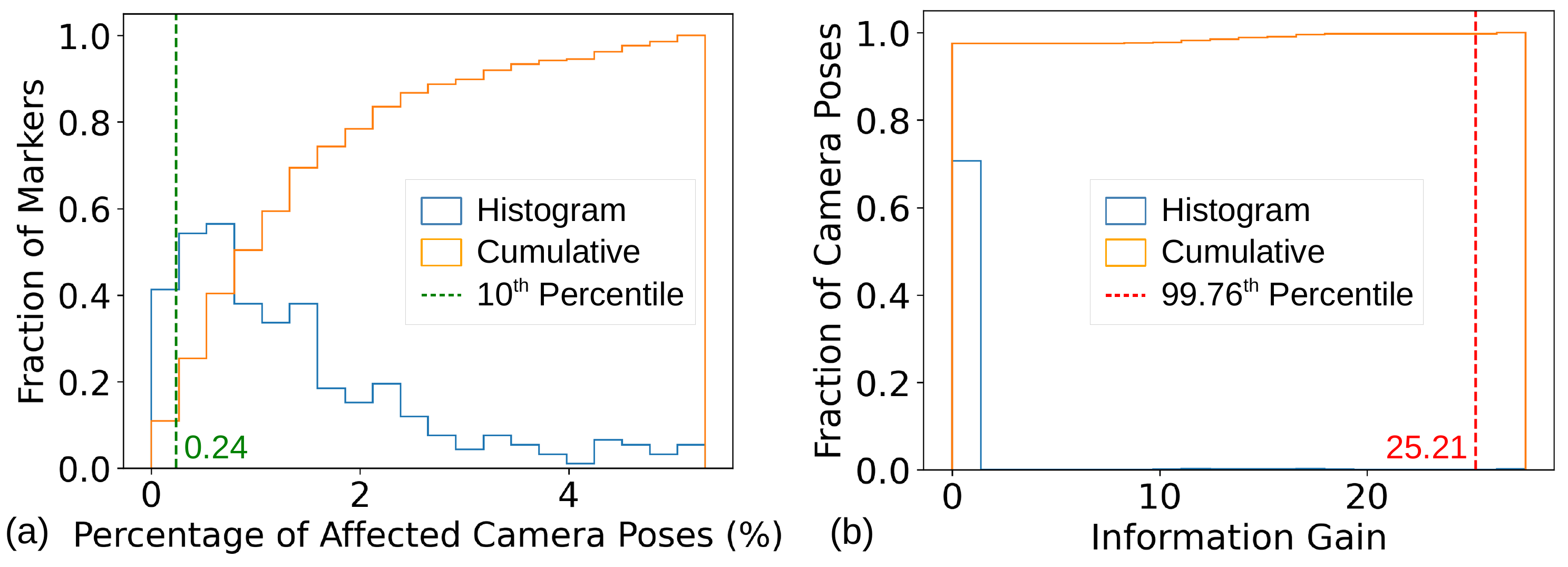}
    \caption{Histograms for the HM3D apartment model: (a) percentage of affected camera poses and (b) information gains at camera poses yielded by a marker. The most visible $90\%$ markers (i.e., $v=90$) means $10^{th}$ percentile in (a), determining the percentile $q=99.76$ by \eqref{eq:q-v-relation}. The $99.76^{th}$ percentile in (b) indicates a localizability gain 25.21 of the marker by \eqref{eq:marker-loc-gain}.}
    \label{fig:pctls}
\end{figure}

\textbf{Analysis of feature similarity of 3D points}: Repetitive structures in scenes cause similar features across RGB images and can result in localizing to a wrong location. To reduce the contribution of repetitive structures to localizability scores, we penalize the localizability score if similar features appear in the FOV of the camera. Specifically, when modeling 3D points with similar features in factor graphs, we set greater uncertainty in noise models of 3D point factors to encode the fact that similar 3D points are ambiguous and less informative. \eqref{eq:point-factor} shows the 3D point factor that formulates the difference between the noisy 3D location $\noisy{\mathbf{p}}$ and true 3D location $\mathbf{p}$ using a Gaussian distribution 
\begin{equation}
    p(\noisy{\mathbf{p}}|\mathbf{p})=\mathcal{N}(\noisy{\mathbf{p}}-\mathbf{p};\mathbf{0},\Sigma_{\mathbf{p}})\label{eq:point-factor}
\end{equation}
where $\Sigma_{\mathbf{p}}$ is the covariance we set for modeling noise.
For example, in the leftmost part of Fig.~\ref{fig:loc-score}, points $\mathbf{p}_1$ and $\mathbf{p}_3$ are visually similar, so we set big covariances in 3D point factors of $\mathbf{p}_1$ and $\mathbf{p}_3$. Informally, factors with big covariances impose loose constraints on the camera pose distribution, leading to lower contributions on the localizability score.

We perform an analysis of feature similarity of 3D points to determine noise models in 3D point factors (i.e., $\Sigma_{\mathbf{p}}$ in \eqref{eq:point-factor}), as shown in the flow chart in Fig.~\ref{fig:loc-score}. The analysis is to count the number of similar 3D points to any 3D point. The resulting covariance $\Sigma_{\mathbf{p}}$ is formulated as
\begin{equation}
    \Sigma_{\mathbf{p}}=(1+n_{\mathbf{p}})\Sigma_0 \label{eq:cov-scaling}
\end{equation}
where $\Sigma_0$ is a base covariance (e.g., $diag(2.5,2.5,2.5)\times 10^{-3}\ m^{2}$ in our experiments) and $n_{\mathbf{p}}$ denotes the number of similar 3D points to the query point $\mathbf{p}$. 3D points observed by all feasible camera poses are filtered to select similar ones of the query point. The selection is determined by two criteria: 1) the selected points have similar descriptors to the query point and 2) the selected points are not too close to the 3D location of the query point. The intuition is that, if two areas in the scene look similar but they are far away from each other, a wrong place recognition would incur a huge localization error.

\textbf{Estimation of camera pose distributions}: We use the Laplace approximation~\cite[Ch. 4.4]{bishop2006pattern} to estimate a Gaussian distribution that approximates the camera pose distribution encountered in the synthetic localization problem. The mean of the Gaussian is the known feasible camera pose so the covariance $\Sigma$ is the only unknown. The covariance can be approximated by an estimated Hessian of the negative logarithm of the camera pose distribution at the mean (see \cite[Sec. 2]{kaess2009covariance} for the estimation of the covariance). Thus the entropy encountered in the synthetic localization problem can be approximated by 
\begin{equation}
H(p(C|\cdot)) \approx \frac{1}{2}\ln |\Sigma| +\frac{d}{2}(1+\ln(2\pi))
\end{equation}
where the dimensionality $d$ is 6 for 6DOF poses.

\subsubsection{The greedy algorithm\protect\footnotemark}\label{sec:greedy}
The algorithm sequentially selects $k$ poses from feasible marker poses $\markerSet$ (see Algorithm~\ref{algo:imp}). The algorithm executes $k$ loops to search the best $k$ poses. In each loop, we update localizability scores, tentatively place a marker at any feasible marker pose, and compute localizability gains of trial marker placements. The best pose that earns the highest localizability gain will be removed from feasible marker poses and be permanently occupied by a marker. The marker will influence future updates of localizability scores.

\footnotetext{\highlight{Discussion about the complexity of the algorithm and the possibility of generalizing the greedy algorithm is available in Sec. \ref{sup:omp}.}}

We summarize information gains at all feasible camera poses in the scene, using a single scalar quantity that we refer to as localizability gain. Informally, one could think of the localizability gain as the reward for placing an additional marker at a specific position. The localizability gain of any marker placement $\markerVal$ is defined as the $q^{th}$ percentile of information gains \highlight{that marker $\markerVal$ yields at all feasible camera poses $\camSet$}, as seen in
\begin{equation}
g(\markerVal)= \inf\{i\in \R: F_{I}(i) \geq \frac{q}{100}\}, \label{eq:marker-loc-gain}
\end{equation}
where $F_I(\cdot)$ is the cumulative distribution function (CDF) \highlight{after sorting the information gains at all camera poses}
\begin{equation}
\mathcal{I}=\{I(\markerVal, \camVal): \camVal \in \camSet\}.
\end{equation}
The choice of percentile $q\in [0,100]$ is crucial and dependent on environments (i.e., the ground plane). For example, in a large environment where any marker is only visible to a small fraction of feasible camera poses, a low percentile $q$ would likely incur zero localizability gains for all markers since camera poses seeing no markers receive zero information gains and constitute a great portion of the information gain distribution $\mathcal{I}$.

We use an adaptive approach to determine the percentile $q$ before computing the localizability gain. The approach introduces a hyperparameter $v\in[0,100]$ and ensures that the most visible $v$ percent of markers earn nonzero localizability gains. A high $v$ allows more markers, even the ones stuck in corners, to effectively join in the selection of best marker while a low $v$ favors the most visible ones among feasible marker poses. In the ground plane space, for any marker $\markerVal$, we can find a set of affected camera poses $\camSet_{\markerVal}$ that are supposed to see the marker (i.e., nonzero info. gain). We can derive a CDF $F_{P}(p)$ using percentages of affected camera poses for all markers 
\begin{equation}
\mathcal{P}= \left\{ \frac{|\camSet_{\markerVal}|}{|\camSet|}\times 100: \markerVal \in \markerSet \right\}.
\end{equation}
To ensure only the most visible $v$ percent of  markers earn nonzero localizability gains, the percentile $q$ is determined by the $(100-v)^{th}$ percentile in percentages of affected camera poses, as seen in
\begin{equation}
q=100 - \inf\left\{p\in [0,100]: F_{P}(p) \geq \frac{100-v}{100}\right\}.\label{eq:q-v-relation}
\end{equation}
\eqref{eq:q-v-relation} indicates $q$ is a non-decreasing function of $v$. When $v$ approaches 100, $q$ approaches 100 as well so only markers that earn a greater maximum in information gains will be considered in the best marker selection (see \eqref{eq:marker-loc-gain}); when $v$ approaches 0, $q$ approaches 0 as well so the best marker will only be selected from markers that influence large areas. Thus the choice of hyperparameter $v$ can reflect the trade-off between helping the worst single camera pose and influencing the most camera poses.

Fig.~\ref{fig:pctls} shows an example for computing the percentile $q$ and the localizability gain for the marker placement in Fig.~\ref{fig:info-gain-res}. We set $v=90$ as the default setting so the most visible $90\%$ markers receive nonzero localizability gains and are effective best marker candidates. This setting results in a marker placement strategy that tends to support worst camera poses instead of area coverage, as shown in the optimized marker placement for the apartment model in Fig.~\ref{fig:all-model-res}. No markers are placed in the two big rooms on the right of the apartment since (i) camera poses in these rooms already enjoyed good localizability scores (see Fig.~\ref{fig:info-gain-res}a) and (ii) a large hyperparameter $v$ does not emphasize area coverage.
\begin{figure}[t]
    \centering
    \includegraphics[width=0.8\linewidth]{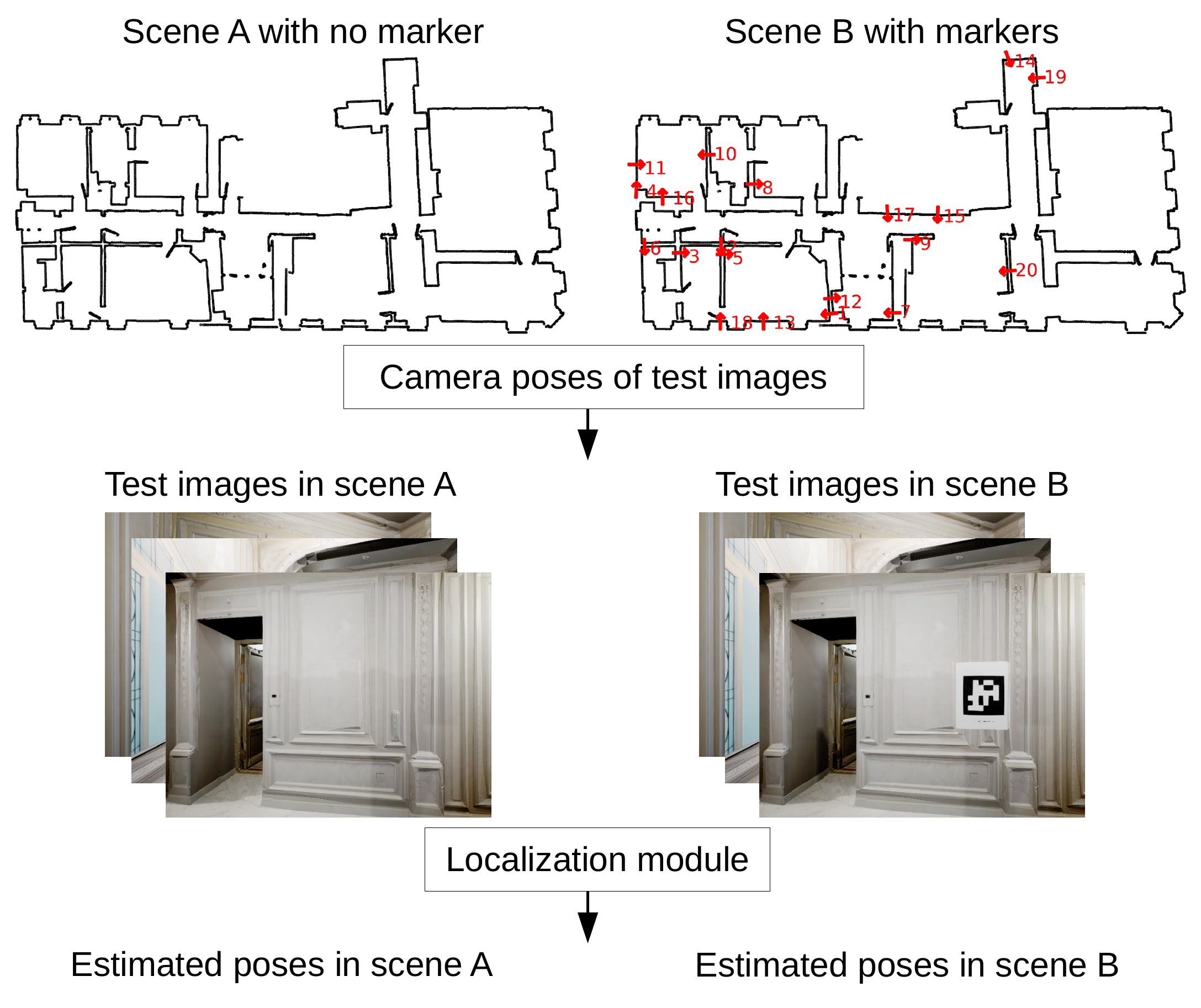}
    \caption{The flowchart of our system for performing camera localization experiments. Scenes with different marker placements share the same set of camera poses for acquiring test images and the same localization module.}
    \label{fig:system}
    \vspace{-5pt}
\end{figure}

\section{Experimental Setup}
\subsection{Implementation}
We implemented all three key elements and Algorithm~\ref{algo:imp} in Sec.~\ref{sec:key-tech} in Python with assistance of a few open source software packages. We used the Unreal Engine 4.27 \cite{unrealengine} and the AirSim library (v1.8.1) \cite{shah2018airsim} to simulate and collect images from 3D models. We used the Open3D library \cite{Zhou2018} to downsample scan points to get candidate marker locations. We used the GTSAM library \cite{gtsam} to create factor graphs and estimate covariances in Gaussian approximations of camera pose distributions. \highlight{The SIFT feature~\cite{Lowe04ijcv} was used throughout our experiments.} 

Additionally, we implemented a simulation system for testing marker placement algorithms and a camera localization module for estimating camera poses of test images. Fig.~\ref{fig:system} presents a flowchart of the system. The system adds markers to a scene model at positions planned by marker placement algorithms and then generates test images from the same set of camera poses for different marker placements for the fairness in comparison. We stress three advantages of the simulation system over real world pipelines for performing camera localization experiments: 1) reproducible data collection by other researchers for future development of marker placement algorithms, 2) a large number of test images that cover the scene, 3) consistent camera poses for generating test images in scenes with different marker placements.

\subsection{Evaluation}
\subsubsection{Methods for comparison} We compare our algorithm OMP with 1) no marker placement, 2) random marker placements, \highlight{3) uniform marker placements, and 4) markers placed by a human}. Random marker placements refer to uniformly weighted samples from feasible marker poses. \highlight{Uniform placements distribute the markers roughly uniformly along the perimeter
of the environment (see \cite{meyer2011using} for details).} We generated 5 versions of random and uniform placements for each scene and all placements were manually inspected in scene models to ensure reasonable quality. The comparison with humans is only conducted in the real experiment. The human prioritizes centers in less textured areas.

\subsubsection{Scenes} The method comparison is performed on four scenes: apartment, studio, office, and \highlight{ room}, as seen in Fig.~\ref{fig:all-model-res}. The first two are pre-built dense maps of realworld spaces, provided by the Habitat-Matterport 3D (HM3D) Research Dataset \cite{ramakrishnan2021hm3d}, while the third model is an Unreal Engine simulation environment that resembles typical realworld offices\protect\footnotemark. The first three are for simulated experiments. \highlight{The last one is a motion capture room at MIT for the real experiment (see Sec. \ref{sup:real-world}). The textured mesh of the room was created by fusing RGB-D images from groundtruth poses, using the volumetric fusion \cite{curless1996volumetric} and marching-cubes algorithms and the screened Poisson surface reconstruction \cite{kazhdan2013screened}.} Table~\ref{model-spec} lists specifics of these models. 

\footnotetext{The serial number of the apartment model is \href{https://aihabitat.org/datasets/hm3d/00770-NBg5UqG3di3/index.html}{00770-NBg5UqG3di3} in the HM3D dataset and that of the studio model is \href{https://aihabitat.org/datasets/hm3d/00254-YMNvYDhK8mB/index.html}{00254-YMNvYDhK8mB}. \highlight{We inspected all scenes in the dataset and chose these two as representatives of medium and large scenes with textureless areas and potential perceptual aliasing.} The office model is the \href{https://www.unrealengine.com/marketplace/en-US/product/threedee-office}{ThreeDee Office} project in the Unreal Engine Marketplace.}

\begin{table}[t]
\caption{Specifics of scenes}
\label{model-spec}
\begin{tabularx}{\linewidth}{@{} l *{3}{C} c @{}}
\toprule
Model  
& Area ($m^2$) & $\#$ of map images & $\#$ of test images\\ 
\midrule
Apartment   & 339.3      & 10856          & 10000  \\ 
Studio & 149.6       & 2832        & 3000  \\ 
Office       & 108.3       & 1768          & 2000  \\  
Room       & 21.0       & 250          & 200\\
\bottomrule
\end{tabularx}
\end{table}

\setlength\floatsep{4pt}
\begin{figure}[t]
    \centering
    \includegraphics[width=0.9\linewidth]{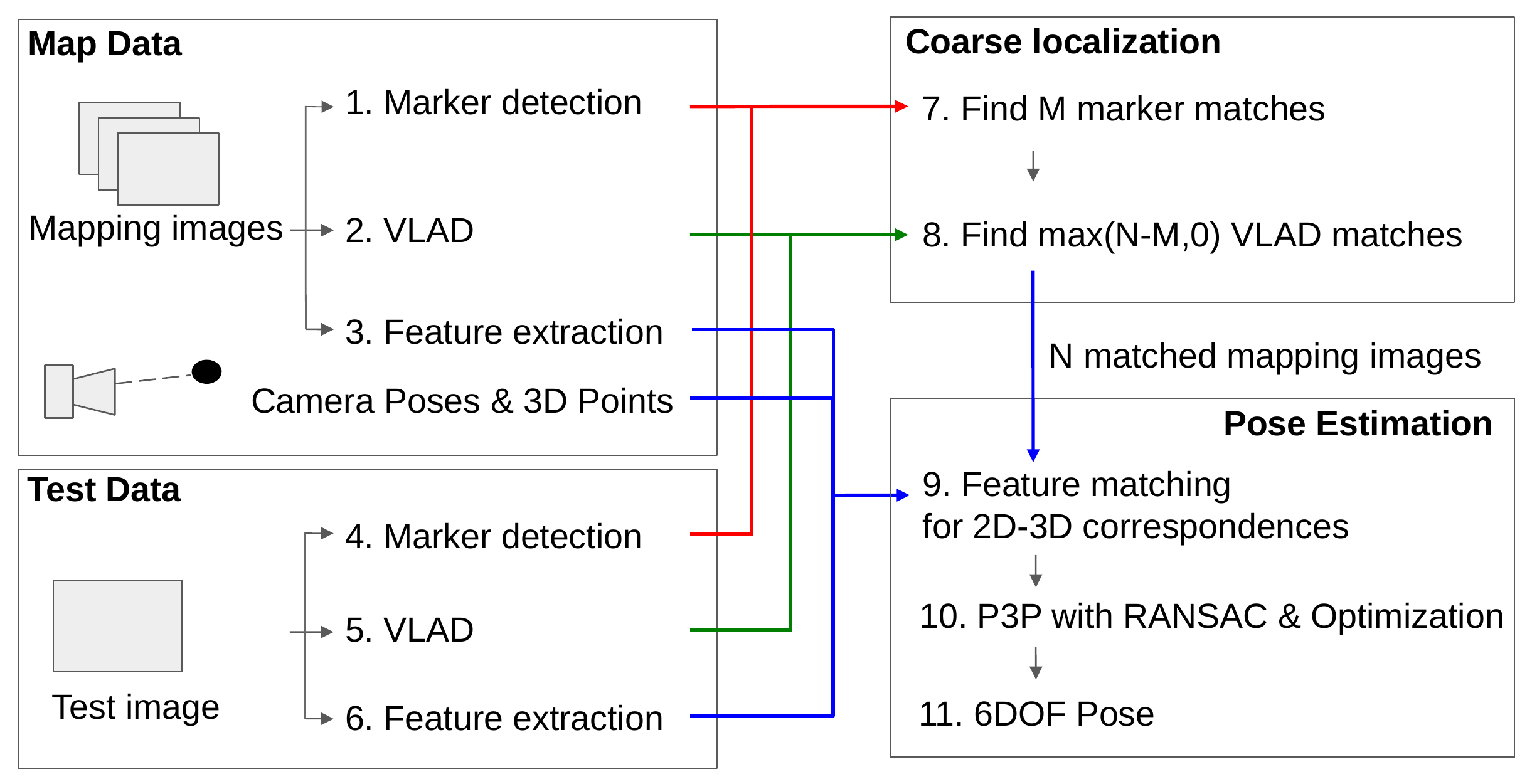}
    \caption{The localization module using fiducial marker detection. The numbers indicate the order of different operations.}
    \label{fig:loc-module}
\end{figure}

\subsubsection{The localization module} Fig.~\ref{fig:loc-module} presents the flowchart of our localization module. The localization module is similar to standard approaches \cite{sarlin2019coarse} but with an extra function of fiducial marker detection, provided by the AprilTag library \cite{olson2011apriltag}. The tag detection and VLAD descriptors \cite{jegou2011aggregating} were sequentially employed to find matched images in the map data. Camera poses were estimated using P3P~\cite{gao2003complete} with RANSAC~\cite{fischler1981random} followed by Levenberg-Marquardt optimization~\cite{opencv_library}. The rotation error $\delta_{\rot}$ is defined as the angular distance between the estimated rotation matrix $\est{\rot}$ and the groundtruth rotation $\TrueRot$ while the translation error $\delta_{\tran}$ is defined as the Euclidean distance between the estimated translation $\est{\tran}$ and the groundtruth translation $\true{\tran}$, as seen in
\begin{align}
\delta_{\rot}&=\big| \text{arccos}\bigl( \frac{\tr({\est{\rot}\transpose \TrueRot})-1}{2} \bigr) \big|,\\
\delta_{\tran}&=\big\| \est{\tran} - \TrueTran \big\|_{2}.
\end{align}

\subsubsection{The map and test data}\label{sec:test-data} \highlight{In simulated experiments, the camera was set to a FOV of 90 degrees and a range of 10 meters (RGB res. 600$\times$450, depth res. 300$\times$225).} The camera poses for collecting the map data are the same as the feasible camera poses in the ground plane space. The camera poses for collecting test images are sampled from the feasible camera poses with weights and then perturbed by translation and rotation noises that are subject to a uniform distribution in $[-0.5, 0.5]$. \highlight{We intend to sample more densely from the difficult areas, which are of our interest}, so the weights in the sampling correlate with localizability scores for generating more test images around low-scoring camera poses\footnote{\label{note1}Results on test images uniformly sampled are available in Sec. \ref{sup:unif}.}. Let $\locSet=\{\locVal(\camVal):\camVal \in \camSet \}$ be the set of localizability scores of feasible camera poses in the ground plane space with no markers. The weights are defined as
\begin{equation}
\weights = \{ 2\locVal^{\star} - \overline{l} - \locVal(\camVal):\camVal \in \camSet  \},
\end{equation}
where $\locVal^{\star}$ is the maximal score in $\locSet$ and $\overline{l}$ is the mean of all scores. Thus all weights will be non-negative and a lower score incurs a greater weight. \highlight{In the real experiment, we used the Realsense L515 camera for RGB-D data (image res. 1280$\times$720) and the OptiTrack system for groundtruth poses. The map and test data were sampled along two lawn-mower paths around feasible camera poses (see Sec. \ref{sup:real-world}).}
\section{Results}
We present two sections of results. In the first section, we present results comparing different marker placement methods. Next, we show a parameter study about factors that can affect our algorithm and the localization performance. The main metric we analyze is the recall, which is defined as the percent of test images localized within certain thresholds of errors: (5 cm, 5 deg) for simulated experiments and \highlight{(30 cm, 10 deg) for the real experiment considering errors in the dense map, sensor noise, and large textureless areas.} The default hyperparameter $v$ is 90.

\subsection{Comparison of Marker Placement Methods}
\begin{figure*}[t!]
    \centering
    \includegraphics[width=0.9\linewidth]{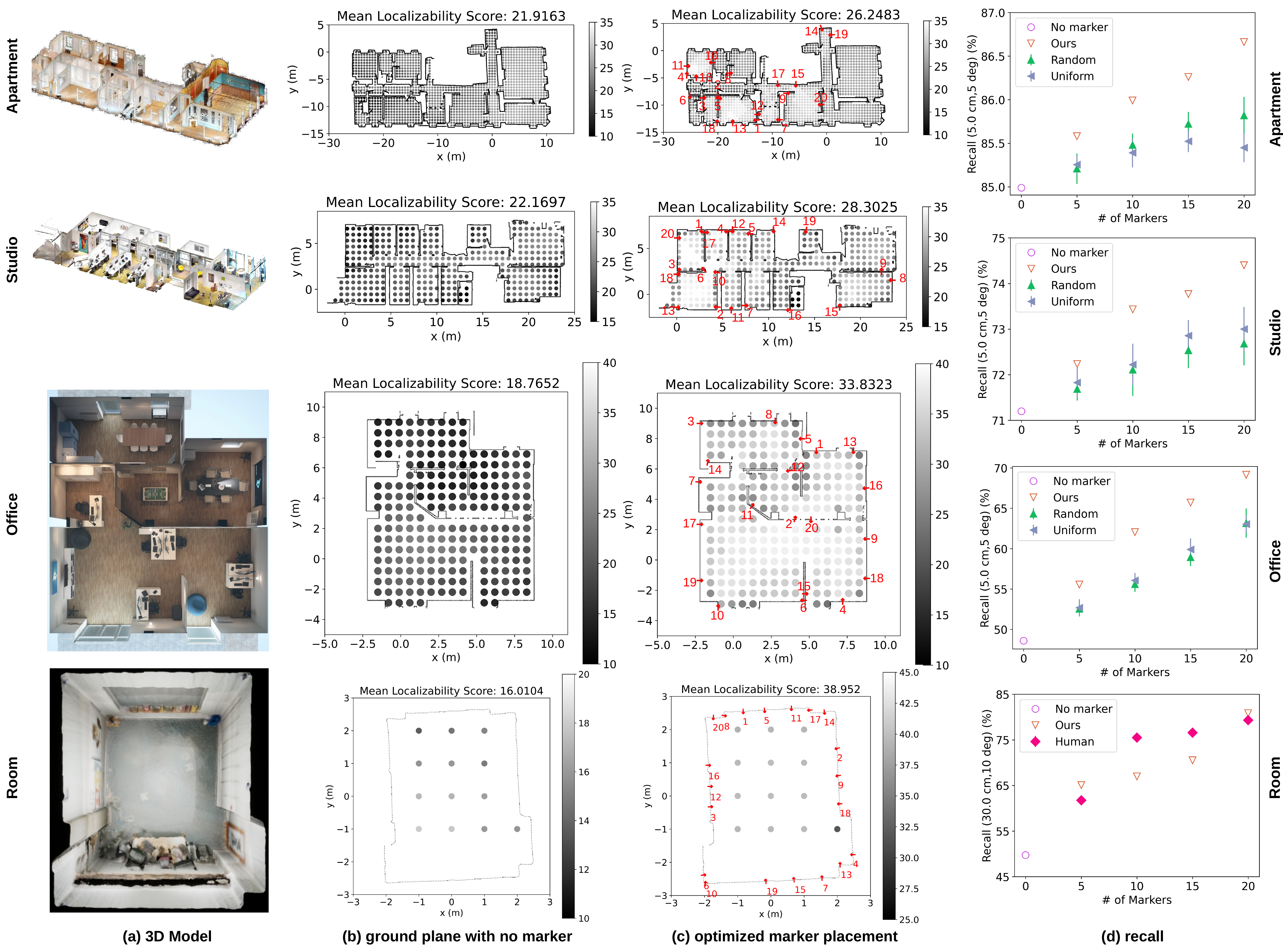}
    \caption{Results for all scenes: (a) 3D models, (b) ground plane space with no markers where darker dots indicate lower localizability scores, (c) optimized marker placements where the red arrows represent optimized marker placements and the numbers beside the arrows indicate the order of marker placements, and (d) the recall in camera localization experiments. We exclude camera poses near the bottom of the room where a table occupies.}
    \label{fig:all-model-res}
\end{figure*}

\textbf{As optimized marker placements in Fig.~\ref{fig:all-model-res} show, our algorithm focuses on placing markers around low-scoring areas and improves mean localizability scores by a large margin}. For example, the largest room in the studio model only receives a single marker (marker 9 on the top right of the studio) since the room already possesses good localizability scores even with no markers.

\textbf{Optimized marker placements consistently outperform no marker placement, random placements, and uniform placements on the recall.} After placing 20 markers, our algorithm improves the recall by over 1.5 percentages for the apartment model, 3.0 percentages for the studio model, 20.0 percentages for the office model, and over 20.0 percentages for the room scene. Note that the area of the apartment model is very big and the model has attained a high recall $85\%$ with no assistance of markers so the increment of recall for the apartment model was expected to be lower than that for the other models. \highlight{The real experiment in the room scene shows that our algorithm is on par with markers placed by a human. Although our experiment demonstrates the efficacy of optimizing marker placements in 3D models for realworld applications, we emphasize that the efficacy relies on the similarity between rendered and real images. Vision features in rendered images can be affected by many factors including mesh quality and lighting. For example, we covered the glass door in the room by a well-textured poster to reduce the difficulty in 3D reconstruction. In addition, if one has quality real RGB-D data at feasible camera poses, the textured mesh is not needed for using our marker placement algorithm.}

\begin{table}
\vspace{+10pt}
\caption{Parameter study about the hyperparameter $v$, the test data, enabling/disabling marker detection, and enabling/disabling the similarity analysis.}
\label{table:param-study}
\begin{tabularx}{\linewidth}{@{} l *{5}{C} @{}}
\toprule
\multirow{2}{*}{Experiment group} & \multicolumn{5}{c}{Recall of test images with $k$ markers (\%)} \\ \cline{2-6} 
                             & $k=0$ & $5$ & $10$ & $15$ & $20$  \\ \midrule
       $v=90$ (df.)         & \textbf{48.6}    & \textbf{55.5}    & \textbf{62.1}    & \textbf{65.7}   & \textbf{69.2}   \\
       $v=99$                & 48.6    & 55.5    & 60.4    & 64.5   & 67.4   \\
       $v=70$                & 48.6    & 54.8    & 61.1    & 63.2   & 66.6   \\
       $v=50$                & 48.6    & 54.3    & 57.6    & 63.9   & 66.8   \\ \midrule
       Marker detect. on (df.)  & \textbf{48.6}    & \textbf{55.5}    & \textbf{62.1}    & \textbf{65.7}   & \textbf{69.2}   \\
       Marker detect. off       & 48.6    & 55.2    & 60.7    & 64.2   & 67.5   \\ \midrule
       Low-scoring data (df.)  & 48.6    & 55.5    & 62.1    & 65.7   & 69.2   \\
       Unif. test data       & \textbf{57.4}    & \textbf{63.7}    & \textbf{68.4}    & \textbf{72.1}   & \textbf{74.8}   \\
\midrule
       Similarity analysis (df.) & \textbf{48.6}    & \textbf{55.5}    & \textbf{62.1}    & \textbf{65.7}   & \textbf{69.2}   \\
       \highlight{Sim. analysis disabled}      & 48.6    & 55.4    & 61.8    & 65.0   & 67.8   \\
\bottomrule      
\end{tabularx}
\vspace{-15pt}
\end{table}

\subsection{Parameter Study}
We design four experiment groups and change one of the default parameters in each experiment group. The experiment groups are 1) different values of $v$ in the greedy algorithm, 2) enabling/disabling marker detection in the localization module, 3) low-scoring/uniform test data and 4) enabling/disabling the analysis of feature similarity, as seen in Table~\ref{table:param-study}. The default setting is with $v=90$, marker detection enabled, the low-scoring test data that has more test images in low-scoring areas in the ground plane, and the similarity analysis where similar 3D points are downweighted in the localizability score. For the parameter study, we use the office model.

\textbf{Too large or small values of hyperparameter $v$ incur lower improvements of the recall.} As explained in Sec.~\ref{sec:greedy}, lower $v$ favors markers that cover larger areas while greater $v$ tends to stress the worst single camera pose. Table~\ref{table:param-study} shows that the default value ($v=90$) consistently outperforms small value 50 and large value 99, indicating that the default attains a good balance between area coverage and helping the worst cases.

\begin{figure}[t]
	\vspace{-10pt}
    \centering
    \includegraphics[width=1.0\linewidth]{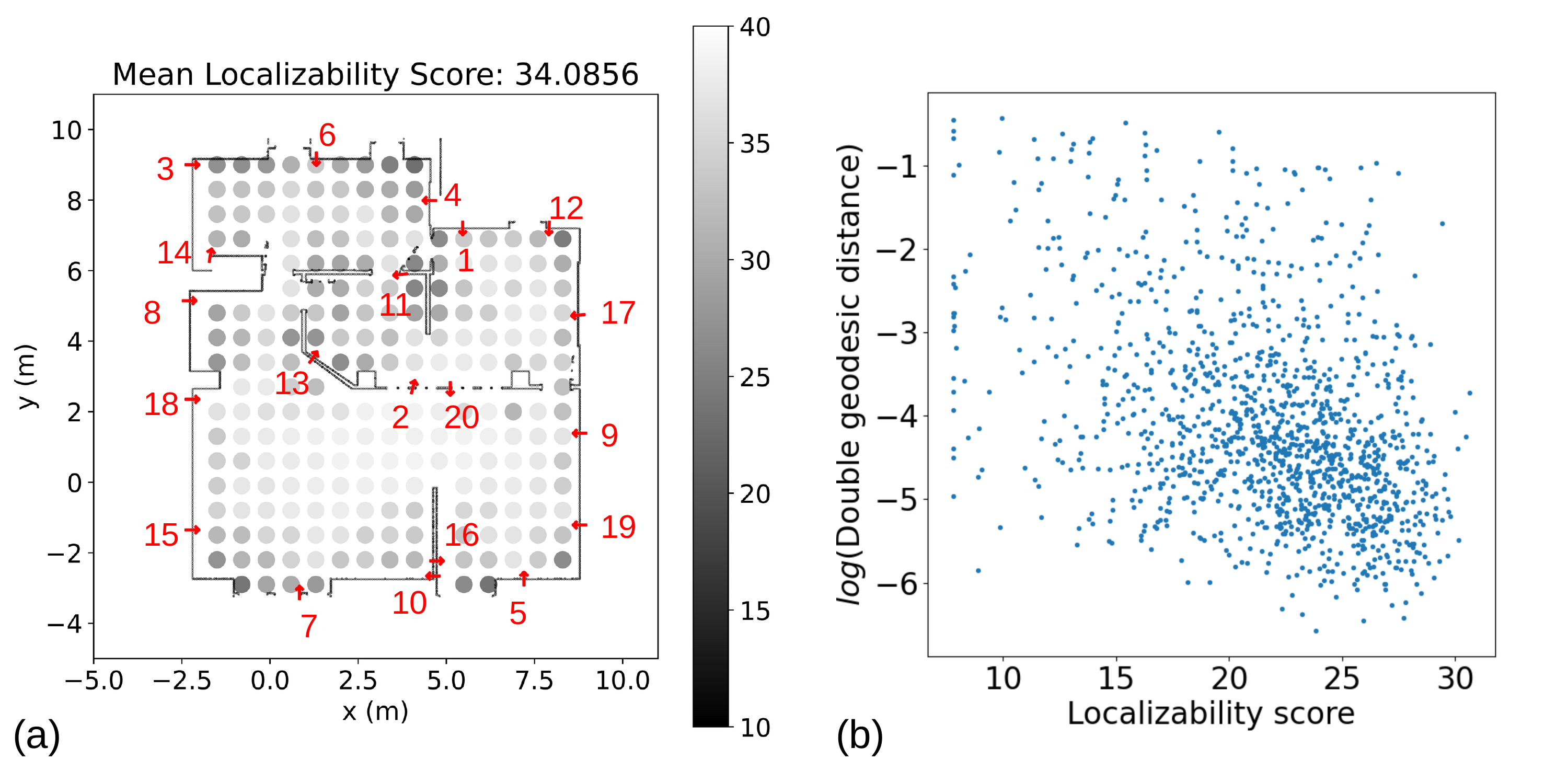}
    \caption{\highlight{Parameter study: (a) the optimized marker placement after disabling the similarity analysis and (b) scatter plot of the localizability score and the log of estimation error of test images. The error is computed as the double Geodesic distance, $\sqrt{\delta_{\rot}^2+\delta_{\tran}^2}$. To avoid outliers, samples are admitted to the plot only if the translation and rotation errors are within $(50 cm, 50 deg)$. The Pearson correlation coefficient and $p$-value for testing non-correlation is $(-0.41,2.4\times10^{-55})$.}}
    \label{fig:param-study}
    	\vspace{-10pt}
\end{figure}

\textbf{The localizability score can be a good indicator of localization errors.} Table~\ref{table:param-study} shows that uniform test samples enjoy greater recall than test samples that stress low-scoring areas by at least 5 percentages. \highlight{Fig.~\ref{fig:param-study}b indicates a statistically significant, negative correlation between the localizability score and the localization error.}

\textbf{Both the visual appearance and decoded label of markers are helpful for localization.} We disable marker detection in the localization module (Fig.~\ref{fig:loc-module}) to investigate its impact on the recall. Table~\ref{table:param-study} shows that markers still improve the recall even though the detector is turned off. The reason is that the visual appearance of markers is still helpful for coarse localization and pose estimation in the localization module.

\highlight{\textbf{Deactivating the analysis of feature similarity decreases the recall.} Fig.~\ref{fig:param-study}a presents the marker placement after disabling the similarity analysis (i.e., no scaling in \eqref{eq:cov-scaling}). The first five markers remain in the same positions as those guided by the similarity analysis (Fig.~\ref{fig:all-model-res}c). Thus the recall does not change significantly until placing 10 markers, as shown in the last group in Table~\ref{table:param-study}. The decrease in the recall with no similarity analysis justifies the efficacy of downweighting similar features in the computation of the localizability score.}
\section{Conclusion and Future Work}
This work provides a promising foundation for optimizing and evaluating marker placement for improved visual localization. Our OMP algorithm defines localizability scores for different areas in the scene and uses a greedy algorithm to find the best marker placements in the sense of increased localizability scores. We applied the OMP algorithm to four scenes and demonstrated that OMP consistently improves camera localization recall compared to random and uniform marker placements. \highlight{We believe that our marker placement approach is also useful for SLAM. However, our approach could be further extended to compute optimal marker placement for specific tasks in SLAM. One potential idea involves extending the localizability score to a trackability score that incorporates uncertainty propagation along a robot path while restricting feasible camera poses to the operating area of the robot.}

The OMP algorithm only considers placing markers in a scene model (i.e., mapped areas in the scene), however, regions in the scene which are challenging for mapping are also likely to be good locations for placing markers. Thus, it would be worth exploring ways to extend the algorithm to prioritize marker placements in regions that are either partially or inadequately mapped. Further research is also needed to compute more accurate localizability scores and explore more efficient optimization methods beyond the greedy algorithm, including: (1) joint optimization of marker poses and sizes, (2) \highlight{extending the single-layer ground plane to multi-layer planes for deploying markers in multi-storey structures}, (3) using non-Gaussian distribution estimation techniques to compute localizability scores, and (4) applying submodular optimization to jointly select multiple best markers together with fewer iterations.


\bibliographystyle{IEEEtran}
\bibliography{ref.bib}

\clearpage

\begin{center}
\large{{\bf Supplementary Material}}
\end{center}
\section{Discussion about the OMP Algorithm}\label{sup:omp}
The crux of computation in Algorithm 1 is evaluating localizability scores. To avoid redundant computation, we update the localizability score of a camera pose only if the newly added marker is covisible to the camera pose. The complexity of Algorithm 1 is $O(|\camSet|+k|\markerSet|\max_{\markerVal} (|\camSet_{\markerVal}|))$ where $O(1)$ is the complexity for evaluating the localizability score of a camera pose. $\max_{\markerVal} (|\camSet_{\markerVal}|)$ denotes the maximal number of covisible camera poses to a marker so it generally increases with the FOV and range of the camera. The first term $|\camSet|$ indicates the cost for initializing localizability scores over all feasible camera poses while $\max_{\markerVal} (|\camSet_{\markerVal}|)$ in the second term bounds the cost for evaluating the localizability gain brought by a marker. For each of the $k$ loops, we evaluate localizability gains of all $|\markerSet|$ markers. Note that the time complexity can be further reduced because it is not necessary to re-evaluate localizability gains for all markers in each loop (lines 8-10 in Algorithm 1). For example, if the covisible camera poses of an unselected marker have not been affected by all selected markers, we do not need to re-compute the localizability gain of that unselected marker.

We discuss the possibility of generalizing the greedy algorithm by re-defining the localizability score. For example, one can use the pose estimation error from a visual localization system (e.g., Fig. 8) to replace the localization score and keep the rest of the algorithm the same. The new marker placement based on the error may enjoy advantages in localization experiments using the same localization system since the marker placement is directly optimized for the system. However, updating the error along with trial marker placements is computationally much more expensive than evaluating the localization score since we need to add markers to the 3D scene model, generate new map and test images, update the map in the localization system, estimate camera poses of test images using the system, and compute the pose estimation error. In contrast, updating the localizability score just needs to re-estimate camera pose distributions, as shown in Fig. 4.

\section{Uniformly sampled test images}\label{sup:unif}
We show recall of uniformly sampled test images in Table \ref{table:unif-test-images}.

\begin{table}[h!]
\vspace{+10pt}
\caption{\highlight{Recall (\%) of test images that are uniformly sampled. Rand. refers to random marker placements and Unif. refers to uniform marker placements.}}
\label{table:unif-test-images}
\begin{tabularx}{\linewidth}{@{} l l *{4}{C} @{}}
\toprule
\multirow{2}{*}{\stackanchor{Scene}{\tiny{(NoMarker)}}} &\multirow{2}{*}{Method} & \multicolumn{4}{c}{Mean$\pm$STD with $k$ markers} \\ \cline{3-6} 
            &                & $k=5$ & $k=10$ & $k=15$ & $k=20$  \\ \midrule
\multirow{3}{*}{\stackanchor{Apt.}{(88.2)}} & Ours    & \textbf{88.6}    & \textbf{88.8}    & \textbf{89.2}   & \textbf{89.5}   \\
& Rand.    & 88.5$\pm$0.1  & 88.8$\pm$0.1 & 89.0$\pm$0.1 & 89.1$\pm$0.2 \\
& Unif.    & 88.4$\pm$0.1    & 88.6$\pm$0.1    & 88.7$\pm$0.1 & 88.9$\pm$0.1 \\
\midrule
\multirow{3}{*}{\stackanchor{Studio}{(80.3)}} & Ours    & \textbf{81.4}    & \textbf{82.7}    & \textbf{83.0}   & \textbf{83.1}   \\
& Rand.    & 80.8$\pm$0.3    & 81.4$\pm$0.4    & 81.2$\pm$0.4   & 81.4$\pm$0.7  \\
& Unif.    & 80.9$\pm$0.2    & 81.2$\pm$0.3    & 81.6$\pm$0.5   & 81.8$\pm$0.2\\
\midrule
\multirow{3}{*}{\stackanchor{Office}{(57.4)}} & Ours    & \textbf{63.7}    & \textbf{68.4}    & \textbf{72.1}   & \textbf{74.8}   \\
& Rand.    & 60.7$\pm$0.7    & 63.0$\pm$1.0    & 66.7$\pm$1.5   & 69.1$\pm$1.8   \\
& Unif.    & 60.0$\pm$0.6    & 63.0$\pm$0.7    & 67.6$\pm$0.9   & 69.9$\pm$1.3\\
\bottomrule      
\end{tabularx}
\end{table}

\section{Setups in the Real experiment}\label{sup:real-world}
\begin{figure}[h!]
    \centering
    \includegraphics[width=1.0\linewidth]{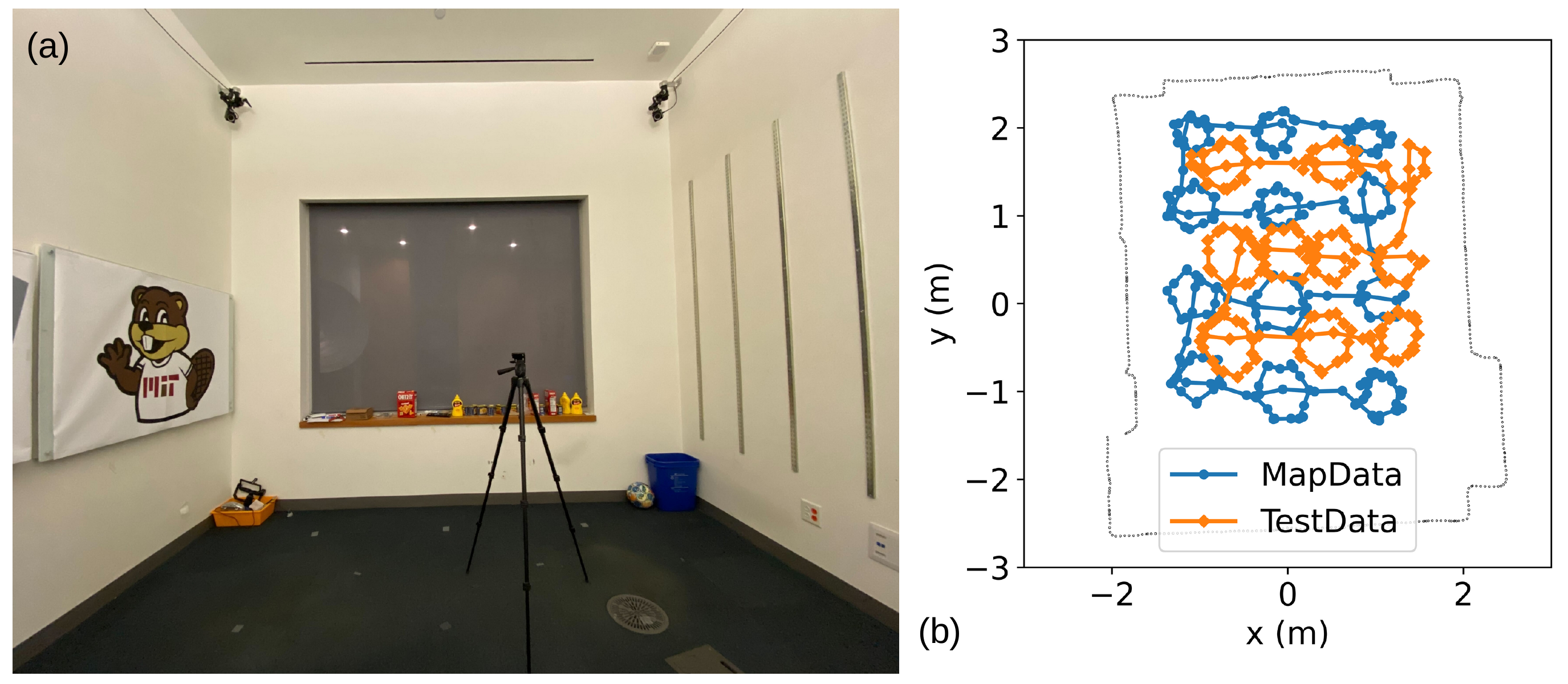}
    \caption{\highlight{Real experiment: (a) the motion capture room and (b) paths for collecting data.}}
    \label{fig:realworld}
\end{figure}

The motion capture room for the real world experiment is shown in Fig.~\ref{fig:realworld}. Fig.~\ref{fig:realworld}b shows the lawn mower paths for collecting the map and test data.

\section{Sensitivity study of marker sizes and positions}\label{sup:sensitivity}
It is quite likely that a user will not be able to place fiducial markers exactly at the positions computed by the OMP algorithm; meanwhile, different users may print fiducial markers with different sizes. Thus we investigate the impact of position deviations and marker sizes on the recall. For the sensitivity study, we used the office model.

\begin{figure}[t!]
    \centering
    \includegraphics[width=1.0\linewidth]{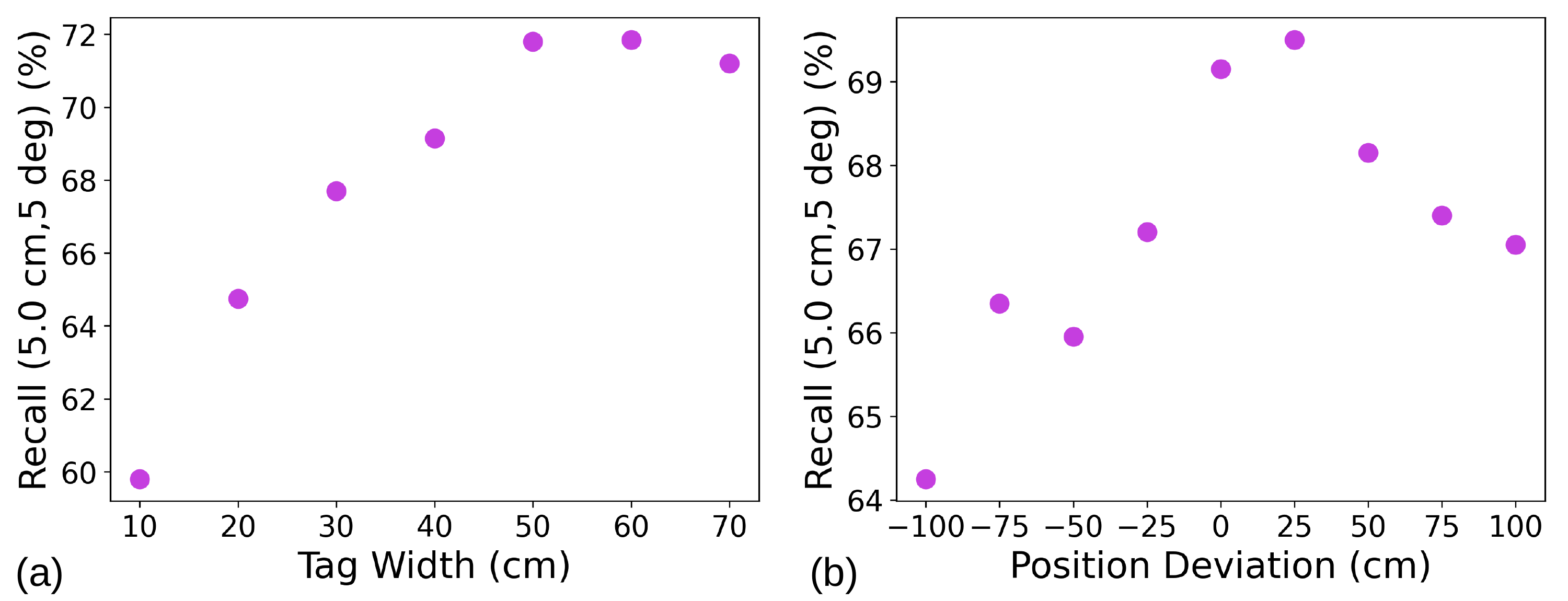}
    \caption{Sensitivity study: (a) re-sizing tags and (b) applying different position deviations to marker poses planned by the OMP algorithm.}
    \label{fig:sens-study}
\end{figure}

\textbf{Enlarging markers up to a certain size keeps increasing the recall.} Fig.~\ref{fig:sens-study}a shows that, under 50 cm, larger tag widths lead to greater recall (note that the threshold 50 cm should correlate with environments). Excessively large sizes can degrade the recall because the markers become too big to be detected from nearby views.

\textbf{Mild position deviations slightly degrade the performance of the optimized marker placement.} All 20 markers planned by the OMP algorithm were moved left or right by certain distances to implement position deviations. Fig.~\ref{fig:sens-study}b shows the recall can decrease by 2\% in the presence of $\pm0.25$ meters position deviations and by 5\% in the presence of $\pm$1 meter position deviations, compared with zero position deviation. However, marker placements with the position deviations still outperform no marker placement by a large margin ($\sim$ 15 percentages in the recall).

\end{document}